# Adaptive Transfer Learning for Plant Phenotyping*

**Jun Wu*[1], Elizabeth A. Ainsworth[2,3,4,5], Sheng Wang[2,3], Kaiyu Guan[2,3,6], Jingrui He[1,7]**

1. *Department of Computer Science, University of Illinois at Urbana-Champaign, Urbana, IL, USA*

2. *Agroecosystem Sustainability Center, Institute for Sustainability, Energy, and Environment, University of Illinois at Urbana-Champaign, Urbana, IL, USA*

3. *College of Agricultural, Consumer and Environmental Sciences, University of Illinois at Urbana-Champaign, Urbana, IL, USA*

4. *Department of Plant Biology, University of Illinois at Urbana-Champaign, Urbana, IL, USA*

5. *USDA ARS Global Change and Photosynthesis Research Unit, Urbana, IL, USA*

6. *National Center for Supercomputing Applications, University of Illinois at Urbana-Champaign, Urbana, IL, USA*

7. *School of Information Sciences, University of Illinois at Urbana-Champaign, Urbana, IL, USA*

**Abstract:** Plant phenotyping (Guo et al. 2021; Pieruschka et al. 2019) focuses on studying the diverse traits of plants related to the plants' growth. To be more specific, by accurately measuring the plant's anatomical, ontogenetical, physiological and biochemical properties, it allows identifying the crucial factors of plants' growth in different environments. One commonly used approach is to predict the plant's traits using hyperspectral reflectance (Yendrek et al. 2017; Wang et al. 2021). However, the data distributions of the hyperspectral reflectance data in plant phenotyping might vary in different environments for different plants. That is, it would be computationally expansive to learn the machine learning models separately for one plant in different environments. To solve this problem, we focus on studying the knowledge transferability of modern machine learning models in plant phenotyping. More specifically, this work aims to answer the following questions. (1) How is the performance of conventional machine learning models, e.g., partial least squares regression (PLSR), Gaussian process regression (GPR) and multi-layer perceptron (MLP), affected by the number of annotated samples for plant phenotyping? (2) Whether could the neural network based transfer learning models improve the performance of plant phenotyping? (3) Could the neural network based transfer learning be improved by using infinite-width hidden layers for plant phenotyping?

To answer the first question, we conduct a variety of experiments on predicting the corn's traits using collected data from (Wang et al. 2021). In this case, partial least squares regression (PLSR) is a commonly used baseline method for plant phenotyping. Besides, we use another two deep learning approaches: multi-layer perceptron (MLP) and Gaussian process regression (GPR). MLP is a class of feedforward artificial neural network, which can universally approximate any measurable function (Hornik et al. 1989). Recently, it is shown (Lee et al. 2018) that deep neural networks with i.i.d. prior over its parameters are equivalent to a Gaussian process, in the limit of infinite network width. Thus, we also consider the Gaussian process regression model (called neural network Gaussian process (NNGP)) derived from deep neural networks. Following (Wang et al. 2021), we test those approaches to predict the leaf photosynthetic traits (e.g., chlorophyll) from leaf hyperspectral reflectance over the optical domain from 400 nm to 2500 nm. When choosing 10% of those data as the training set and others as the test set, we obtain that the results of PLSR, MLP and NNGP are 0.7363|0.6923|0.6666 µg/cm² in RMSE (0.6579|0.6975|0.7195 in $R^2$), respectively. Moreover, with less training data (e.g., 5% as the training set), the results become 1.0058|1.1547|0.6272 in RMSE (0.3662|0.1646|0.7535 in $R^2$), respectively. These results indicate that the performance of conventional machine learning models would significantly degrade when there are only limited training samples. In particular, deep neural networks would be more likely to be over-fitting when only limited training samples are available.

Due to the sensitivity of machine learning models on the number of annotated samples, we then study the neural network based transfer learning models for answering the second question. Transfer learning (Pan et al. 2009) refers to the knowledge transfer from the source task to the target task such that the prediction performance on the target task could be significantly improved as compared to learning from the target task alone. For example, in the context of transfer learning, it can use the simulated directional-directional leaf hyperspectral reflectance (Jay et al. 2016) as the source task, and the real hyperspectral reflectance of crop leaves (Wang et al. 2021) as the target task. It is also notable that one assumption of transfer learning is the relatedness of source and target tasks. If those tasks are not related, it might lead to negative transfer (Wang et al. 2019; Wu et al. 2021). Thus, the selection of source task is important for guaranteeing the success of knowledge transfer across tasks. The crucial procedures of transfer learning include pre-training and fine-tuning. That is, it first pretrains the MLP model using the simulated directional hyperspectral reflectance, and then fine-tune the model using the limited training data from real hyperspectral reflectance of crop leaves. Following our previous setting, when choosing 5% of target data as the training set and others as the test set, we obtain that the result of transfer learning is 0.5375 µg/cm² in RMSE (0.8189 in $R^2$). The results demonstrate that the knowledge from the simulated hyperspectral reflectance can help improve the performance of plant phenotyping.

It is observed from the aforementioned experiments that (1) Gaussian process regression that recovers the deep neural networks with infinite width performs better than the finite-width neural networks, and (2) transfer learning improves the performance of plant phenotyping by leveraging the knowledge from another relevant source task. These results motivate us to think about the third question, i.e., whether the transfer learning approaches can be improved by using infinite-width hidden layers. To this end, we present a Gaussian process based adaptive transfer learning approach (Cao et al. 2010) for plant phenotyping. The key idea is to construct the transfer kernel matrix $K = \begin{bmatrix} K_{ss} & \lambda K_{st} \\ \lambda K_{st} & K_{tt} \end{bmatrix}$ where $K_{ss}$, $K_{st}$, $K_{tt}$ represent the kernel matrix of source-source, source-target and target-target samples, and $\lambda$ measures the relatedness of source and target tasks. In this case, one intuitive idea is to define the base kernel matrix $K_{ss}$, $K_{st}$, $K_{tt}$ using the NNGP kernel (Lee et al. 2018). However, it also introduces additional challenges: (1) the estimate of task relatedness $\lambda$, and (2) scalability on calculating the inverse of the kernel matrix. Specifically, it is very expansive the calculate the kernel matrix from a large number of samples.

Therefore, it is concluded that adaptive transfer learning could help improve the performance of plant phenotyping by leveraging the knowledge from other similar tasks, especially when there is only a limited number of training samples for plant phenotyping. In addition, the knowledge transferability of adaptive transfer learning also allows re-training a new model on different plants or the same plant from different environments efficiently.